\documentclass[journal]{IEEEtran}
\ifCLASSINFOpdf
  \usepackage[pdftex]{graphicx}
  \graphicspath{{./img/}}
  \DeclareGraphicsExtensions{.pdf,.jpeg,.png}
\else
\fi
\usepackage{amsmath}
\usepackage{url}

\usepackage{multirow}

\begin{document}
%
\title{FPGA Implementation of Convolutional Neural Networks with Fixed-Point Calculations}
%
%
%

\author{Roman~A.~Solovyev,
		Alexandr~A.~Kalinin,
        Alexander~G.~Kustov,
        Dmitry~V.~Telpukhov,
        and~Vladimir~S.~Ruhlov
\thanks{R. A. Solovyev, A. G. Kustov, D. V. Telpukhov, and V. S. Ruhlov are with the Institute for Design Problems in Microelectronics of Russian Academy of Sciences (IPPM RAS), Moscow 124365, Russian Federation.}
\thanks{A. A. Kalinin is with the Department
of Computational Medicine and Bioinformatics, University of Michigan, Ann Arbor,
MI, 48104 USA}
}

\maketitle

\begin{abstract}
Neural network-based methods for image processing are becoming widely used in practical applications. Modern neural networks are computationally expensive and require specialized hardware, such as graphics processing units. Since such hardware is not always available in real life applications, there is a compelling need for the design of neural networks for mobile devices. Mobile neural networks typically have reduced number of parameters and require a relatively small number of arithmetic operations. However, they usually still are executed at the software level and use floating-point calculations. The use of mobile networks without further optimization may not provide sufficient performance when high processing speed is required, for example, in real-time video processing (30 frames per second). In this study, we suggest optimizations to speed up computations in order to efficiently use already trained neural networks on a mobile device. Specifically, we propose an approach for speeding up neural networks by moving computation from software to hardware and by using fixed-point calculations instead of floating-point. We propose a number of methods for neural network architecture design to improve the performance with fixed-point calculations. We also show an example of how existing datasets can be modified and adapted for the recognition task in hand. Finally, we present the design and the implementation of a floating-point gate array-based device to solve the practical problem of real-time handwritten digit classification from mobile camera video feed.
\end{abstract}


%
\IEEEpeerreviewmaketitle

\section{Introduction}
%
%
%
%
\IEEEPARstart{R}{ecent} research in artificial neural networks has demonstrated their ability to perform well on a wide range of tasks including image, audio, and video processing and analysis in many domains \cite{lecun2015deep,ching2018opportunities}.
In some applications, deep neural networks have been shown to outperform conventional machine learning methods and even human experts \cite{he2015delving,szegedy2015cvpr}.
Most of the modern neural network architectures for computer vision
include convolutional layers and thus are called convolutional neural networks (CNNs). They have high computational requirements such that even modern central processing units (CPUs) are often not fast enough and specialized hardware, such as graphics processing units (GPUs), is needed \cite{lecun2015deep}.
However, there a compelling need for the use of deep convolutional neural networks on mobile devices and in embedded systems.
This is particularly important for video processing in, for example, autonomous cars and medical devices \cite{shvets2018automatic,shvets2018angiodysplasia}, which demand capabilities of high-accuracy and real-time object recognition.

Following properties of many modern high-performing CNN architectures make their hardware implementation feasible:
\begin{itemize}
\item high regularity: all commonly used layers have similar structure (\texttt{Conv3x3}, \texttt{Conv1x1}, \texttt{MaxPooling}, \texttt{FullyConnected}, \texttt{GlobalAvgPooling});
\item typically small size of convolutional filters: $3\times3$;
\item \texttt{ReLU} activation function (comparison of the value with zero): easier to compute compared to previously used \texttt{Sigmoid} and \texttt{Tanh} functions.
\end{itemize}
Due to high regularity, size of the network can be easily varied, for example, by changing the number of convolutional blocks. In the case of field programmable gate arrays (FPGAs), this allows to program the network on different types of FPGAs, providing different processing speed. For example, implementation of higher number of convolutional blocks on an FPGA can directly lead to a speed-up in processing.

Related direction in neural network research considers adapting them for the use on mobile devices, for example, see MobileNet \cite{howard2017mobilenets} and SqueezeNet \cite{iandola2016squeezenet}. Mobile networks typically have reduced number of weights and require relatively small number of arithmetic operations. However, they are still executed at the software level and use floating-point calculations. For some tasks such as real-time video analysis that requires processing of 30 frames per second mobile networks still can be not fast enough without further optimization.

In order to use an already trained neural network in a mobile device, a set of optimizations can be used to speed up computation. There exist a number of approaches to do so, including weight compression \cite{han2015deep} or computation using low-bit data representations \cite{sato2017tpu}.

Since hardware requirements for neural networks keep increasing, there is a need for design and development of specialized hardware block for the use in ASIC and FPGA. The speed up can be achieved by following:
\begin{itemize}
\item hardware implementation of the convolution operation, which is faster than software convolution;
\item using fixed-point arithmetic instead of floating-point calculations;
\item reducing the network size while preserving the performance;
\item modifying the structure of a network architecture while preserving the same level of performance and decreasing the footprint of the hardware implementation and saved weights.
\end{itemize}

For example, Qiu J. \textit{et al.} \cite{qiu2016fpga} proposed an FPGA implementation of pre-trained deep neural networks from VGG-family \cite{simonyan2014very}. They used dynamic-precision quantization with 4\/8-bit data representation and singular vector decomposition to reduce the size of fully-connected layers, which led to smaller number of weights that had to be passed from the device the external memory. Zhang C. \textit{et al.} \cite{zhang2015fpga} quantitatively analyzed computing throughput and required memory bandwidth for various CNNs using optimization techniques, such as loop tiling and transformation. This allowed their implementation to achieve a peak performance of 61.62 GFLOPS.
Related approach is suggested in \cite{han2016eie}, which allowed to reduce power consumption by the compression of network weights. Higher level solution is proposed in \cite{suda2016opencl}, which considers the use of the OpenGL compiler for deep networks, such as AlexNet and VGG. Duarte \textit{et al.} \cite{duarte2018fast} have recently suggested the protocol for automatic conversion of neural network implementations in high-level programming language to intermediate format (HLS) and then into FPGA implementation. However, their work is mostly focused on the implementation of fully-connected layers. 

In this work we propose a design and implementation of FPGA-based CNN with fixed-point calculations that allows to achieve the exact performance of the corresponding software implementation on the live handwritten digit recognition problem. Due to the reduced number of parameters we avoid common issues with memory bandwidth. Suggested method can be implemented on a very basic FPGAs, but also is scalable for the use on FPGAs with large number of logical cells. Additionally, we demonstrate how existing open datasets can be modified in order to better adapt them for real-life applications. Finally, in order to promote the reproducibility of results, facilitate open-scientific development, and enable collaborative validation we make our source code, documentation, and all results from this study available online.

 


\begin{figure}[!t]
\centering
\includegraphics[width=\linewidth]{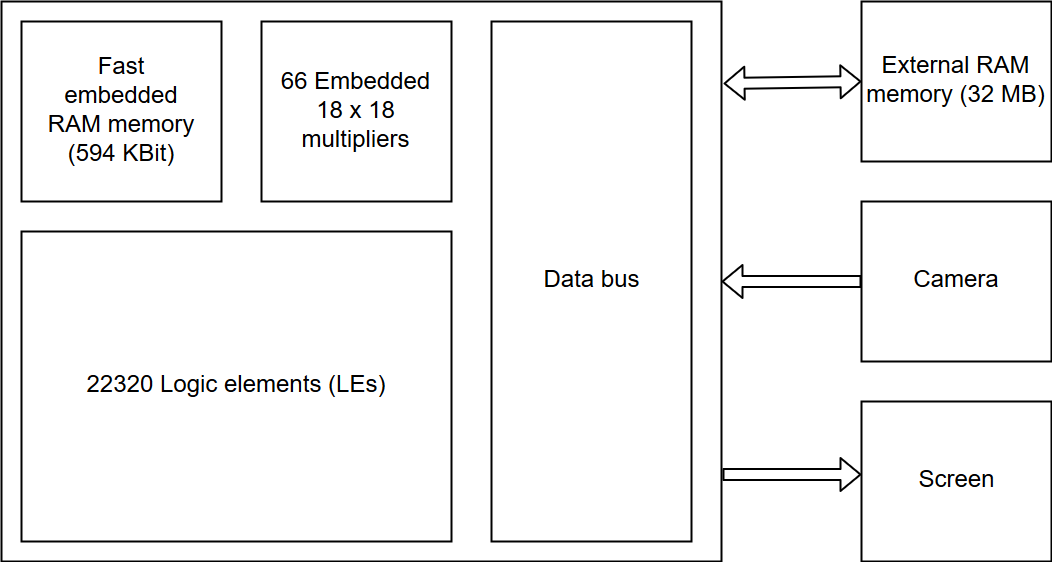}
\caption{DE0-Nano development board and external devices.}
\label{fig:board}
\end{figure}

\section{Methods}
\subsection{Implementation requirements}
To demonstrate our approach, we implement a solution for the problem of recognizing handwritten digits received from a camera in real time. The results are displayed on an electronic LED screen. The minimal speed of digit recognition should exceed 30 FPS, that is, a neural network should be able to process a single image in 33ms. The resulting hardware implementation should be ready for transfer to separate custom VLSI device for mass production.

\subsection{Hardware specifications}
We use the compact development board DE0-Nano \cite{terasicnano} due to the following reasons:
\begin{itemize}
\item Intel (Altera) FPGA is installed on this board, which is mass-produced and cheap;
\item Cyclone IV FPGA has rather low performance and small number of logic cells, assuming increased performance if re-implemented with most of other modern FPGAs;
\item it makes connecting peripherals, such as camera and touchscreen, easier;
\item the board itself has 32 MB of RAM, which can be used to store weights of a neural network.
\end{itemize}
The general scheme of the board and external devices is shown in Figure \ref{fig:board}.

\begin{figure}[!t]
\centering
\includegraphics[width=\linewidth]{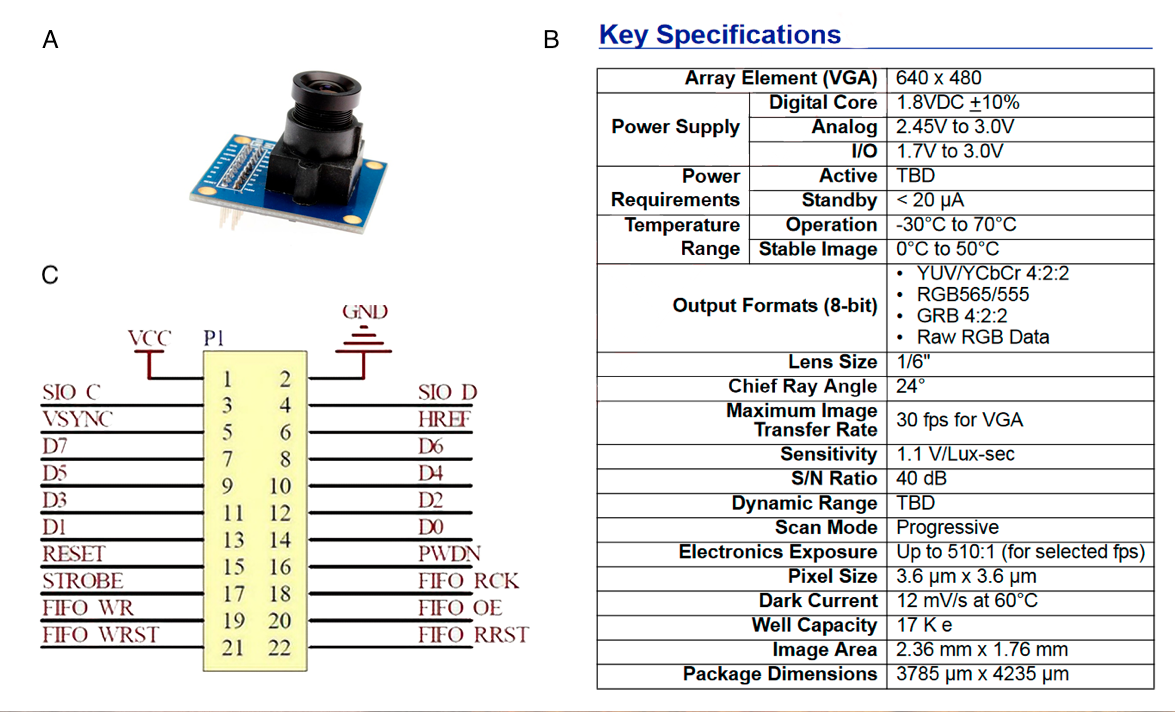}
\caption{Camera module characteristics: (A) physical appearance; (B) technical specifications; (C) pinout scheme.}
\label{fig:camera}
\end{figure}

\begin{figure}[!t]
\centering
\includegraphics[width=\linewidth]{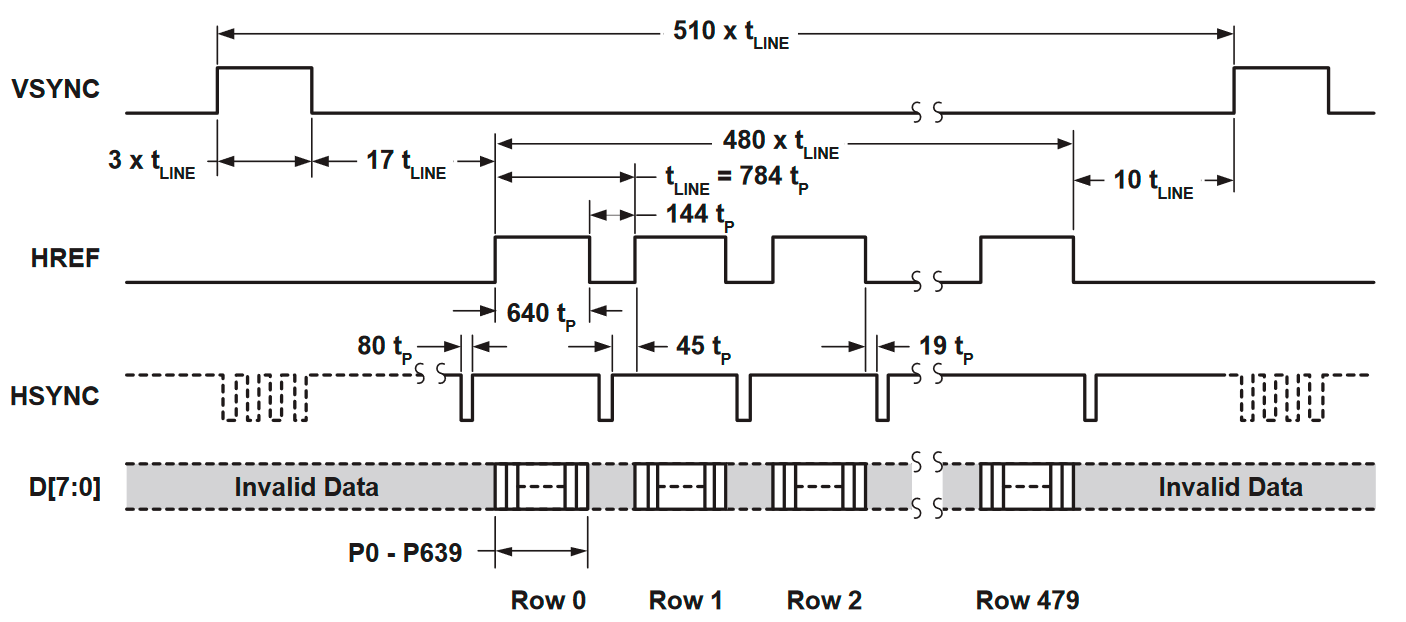}
\caption{Camera operation waveforms.}
\label{fig:waveforms}
\end{figure}

\begin{figure}[!t]
\centering
\includegraphics[width=\linewidth]{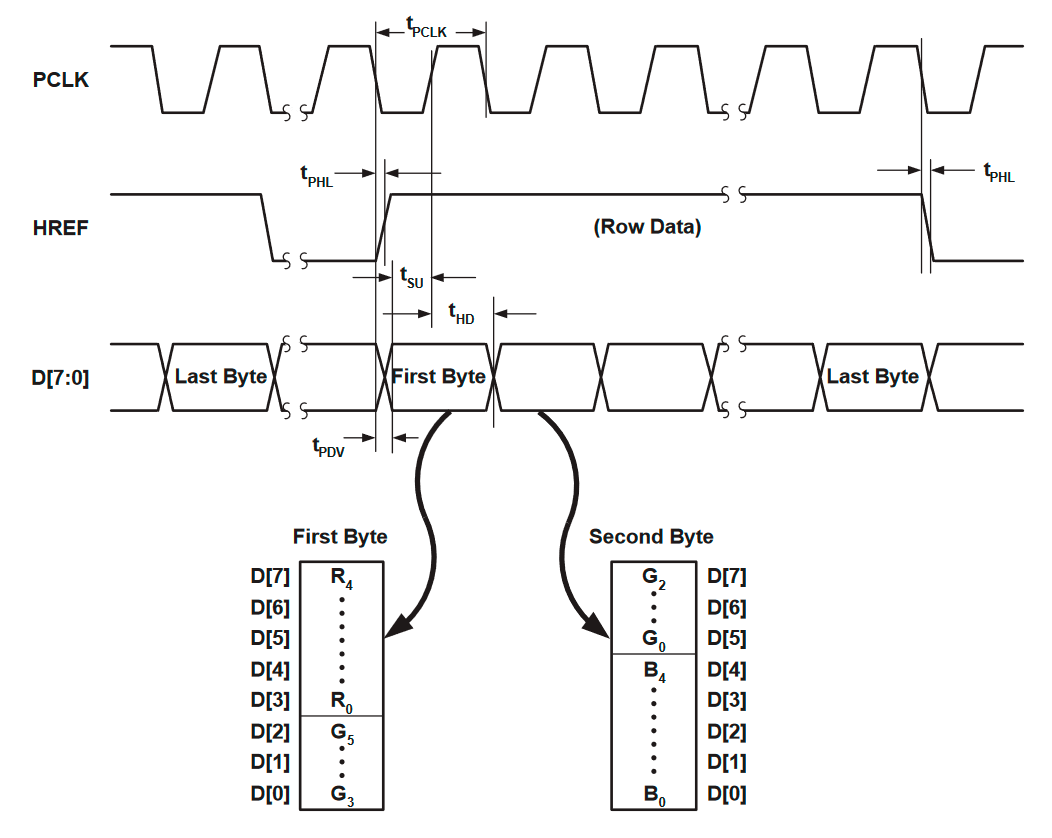}
\caption{Transmitting color data from the camera module.}
\label{fig:transmissioncolor}
\end{figure}

OV7670 camera module (Fig. \ref{fig:camera}) is chosen for image acquisition due to the high quality/price ratio. In this application, high resolution video is not required, since every image is reduced to the size of $28\times28$ pixels and converted to grayscale. The camera module also has a simple connection mechanism (Fig. \ref{fig:camera}C). Only 7 pins are used to interact with the user. Data are transmitted via 8-bit bus using synchronization strobes VSYNC and HREF. SIOC clock signal and SIOD data signal are used to adjust camera parameters. PWDN signal is used to turn on the camera, and RESET for the reset operation. Remaining pins are used for FIFO on camera board. Camera operation waveforms are shown in Fig. \ref{fig:waveforms}. Color data transmission takes 2 clock cycles. Data packing is presented in Fig. \ref{fig:transmissioncolor}.

\begin{figure}[!t]
\centering
\includegraphics[width=\linewidth]{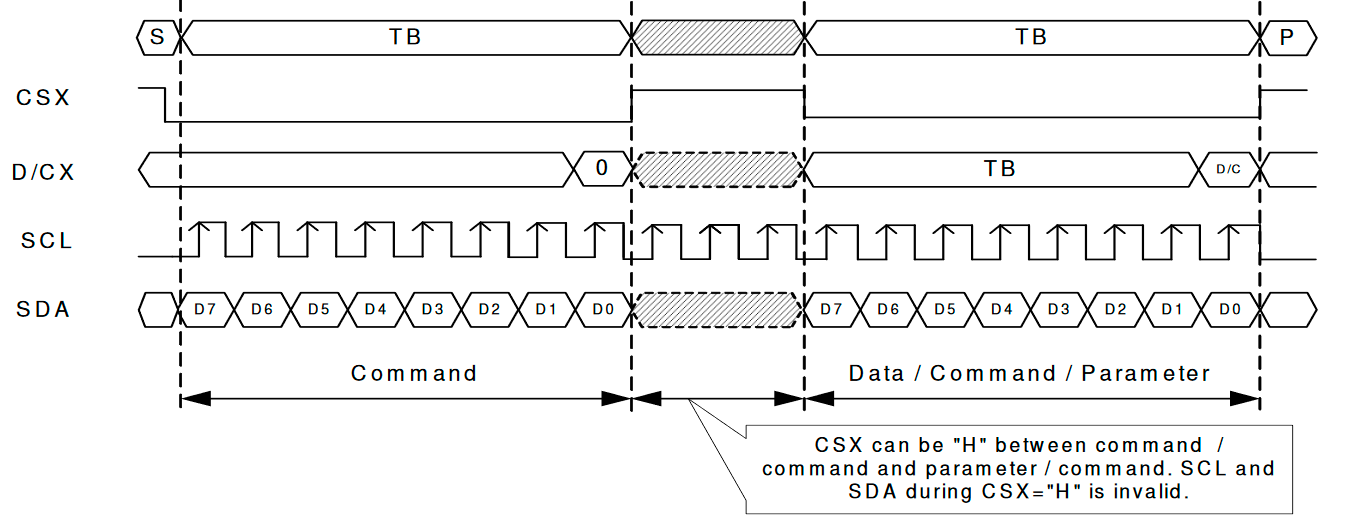}
\caption{Data transmission via SPI interface.}
\label{fig:transmissionspi}
\end{figure}

Display module with $320\times240$ resolution TFT screen is chosen as the output device. Display module driven by microcontroller is equipped by 2.4 inches color touchscreen (18-bit color, 262,144 color variations). It also has backlight and is convenient to use due to the large viewing angle. Contrast and dynamic properties of H24TM84A LCD indicator allow displaying video. LCD controller contains RAM buffer that lowers the requirements for the device microcontroller. Display is controlled via serial SPI bus, as shown in Fig. \ref{fig:transmissionspi}.

\begin{figure}[!t]
\centering
\includegraphics[width=\linewidth]{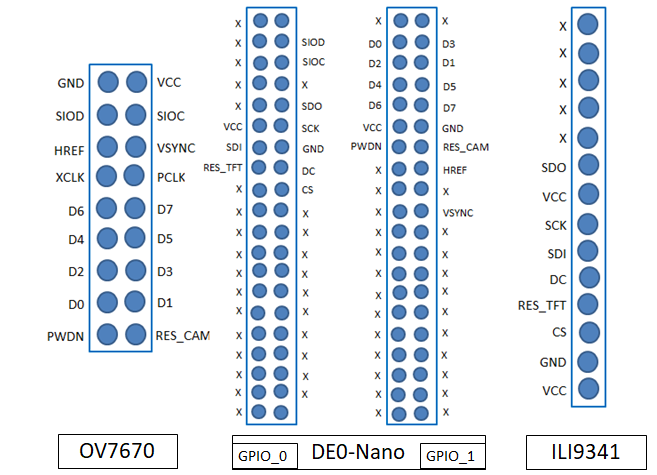}
\caption{Scheme for connecting camera and screen modules to De0-Nano board. Pins with the same name are connected. Pins marked as 'x' remain unconnected.}
\label{fig:connection}
\end{figure}

Final scheme for connecting camera and screen modules to De0-Nano board is shown in Fig. \ref{fig:connection}.

\subsection{Dataset preparation}
\begin{figure}[!t]
\centering
\includegraphics[width=\linewidth]{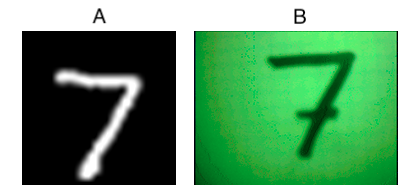}
\caption{Different appearance of (A) an image from the MNIST dataset; and (B) an image from the camera feed.}
\label{fig:mnist}
\end{figure}

The MNIST dataset for handwritten digit recognition \cite{lecun1998mnist} is widely used in the computer vision community. However, it is not well suited for training a neural network in our application, since it differs greatly from the camera images (Fig. \ref{fig:mnist}).

Major differences include:
\begin{itemize}
\item MNIST images are light digits over dark background, opposite to those from the camera feed;
\item camera produces color images, while MNIST is grayscale;
\item size of MNIST image is $28\times28$ pixels, while camera image size is $320\times240$ pixels;
\item unlike centrally placed digits and homogenious background in MNIST images, digits can be shifted and slightly rotated in camera images, sometimes with noise in the background;
\item MNIST does not have a separate class of images without digits.
\end{itemize}

Given that the recognition performance on the MNIST dataset is very high (modern networks recognize numbers with accuracy of more than 99.5\% \cite{wan2013regularization}), we reduce the size of image from camera to $28\times28$ pixels and convert them into grayscale. This helps us to address following problems:
\begin{itemize}
\item there is no significant loss in accuracy, as even in small images digits are still easily recognized by humans;
\item color information is excessive for digit recognition;
\item noisy images from camera can be cleaned by reducing and averaging neighboring pixels.
\end{itemize}

Since image transformation is also performed at hardware level, it is necessary to consider in advance a minimum set of arithmetic functions that can effectively bring image to the desired form. The suggested algorithm for modifying camera images goes as following:

\begin{enumerate}
\item We crop a central part measuring $224\times224$ pixels from a $320\times240$ image, which subsequently allows easy transition to the desired image size, since $224 = 28\times8$.
\item Then, a cropped image part is converted to a grayscale image. Because of the peculiarities of human visual perception \cite{burger2009imageprocessing}, we take weighted, rather than simple, average. To facilitate the conversion at the hardware level, the following formula is used:
\begin{equation}
\label{bw}
    BW = (8*G + 5*R + 3*B) / 16
\end{equation}
Multiplication by 8 and division by 16 are implemented using shifts.
\item Finally, $224\times224$ image is split into $8\times8$ blocks. We calculate average value for each of these blocks, forming a corresponding pixel in $28\times28$ image.
\end{enumerate}
Resulting algorithm is simple and works very fast at the hardware level.

In order to use MNIST images for training a neural network, on-the-fly data augmentation is used. This method implies that during the creation of the next mini-batch for training, a set of different filters is arbitrarily applied to each image. This technique is used to easily increase the dataset size, as well as to bring images to the required form, as in our case.

The following filter set was used for augmenting MNIST images:
\begin{itemize}
\item color inversion;
\item random 10 degrees rotation in both directions;
\item random expansion or reduction of an image by 4-pixels;
\item random variation of image intensity (from 0 to 80);
\item adding random noise from 0\% to 10\%.
\end{itemize}
Optionally, images from camera can be mixed into mini-batches.

\subsection{CNN architecture design}
Despite recent developments in CNN architectures \cite{he2016deep,szegedy2015cvpr}, the essence remaines the same: the input size decreases from layer to layer and the number of filters increases. At the end of a network, a set of characteristics is formed that are fed to the classification layer (or layers), and the output neurons indicate the likelihood that the image belongs to a particular class.

The following set of rules for constructing a neural network architecture is proposed to minimize the total number of stored weights (which is critical for mobile systems) and facilitate the transfer to fixed-point calculations:
\begin{itemize}
\item minimize number of fully connected layers, which consume major part of memory for storing weights;
\item reduce number of filters of each convolution layer as much as possible without degrading the classification performance;
\item stop using bias, which is important when shifting from floating-point to fixed-point, because adding a constant hinders monitoring range of values, and rounding bias error over each layer tends to accumulate;
\item use simple type activation, such as RELU, since other activations, such as sigmoid and tahn, contain division, exponentiation, and other functions that are harder to implement in hardware;
\item minimize number of heterogeneous layers, so that one hardware unit can perform calculations at a large number of flow stages.
\end{itemize}

Before translating the neural network onto hardware, we train it on a prepared dataset and save the software implementation for testing. We create software implementation using Keras with Tensorflow backend, a high-level neural networks API in Python \cite{chollet2015keras}.

\begin{figure}[!t]
\centering
\includegraphics[width=\linewidth]{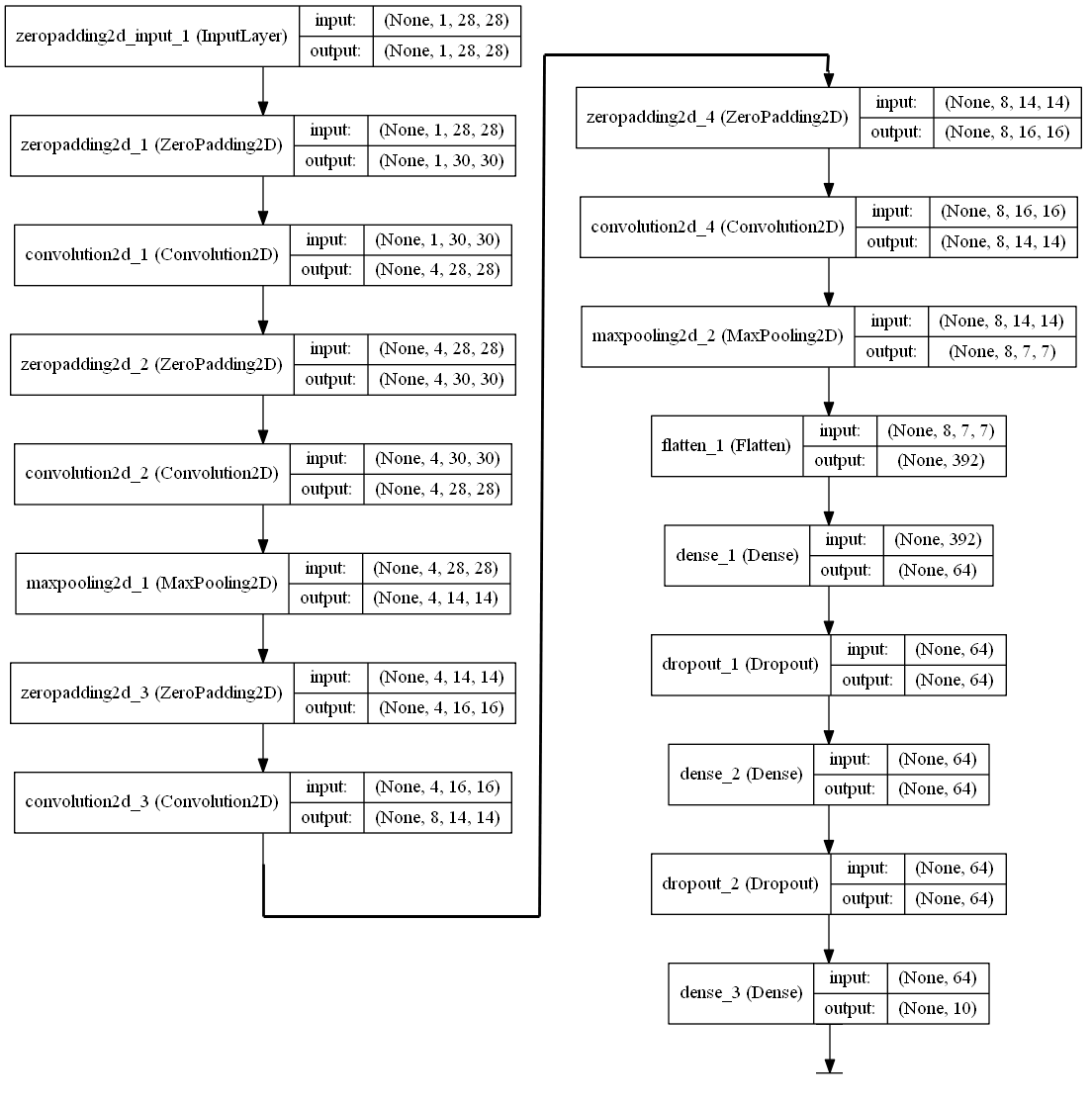}
\caption{VGG Simple neural network architecture.}
\label{fig:vggsimple}
\end{figure}

\begin{figure}[!t]
\centering
\includegraphics[width=\linewidth]{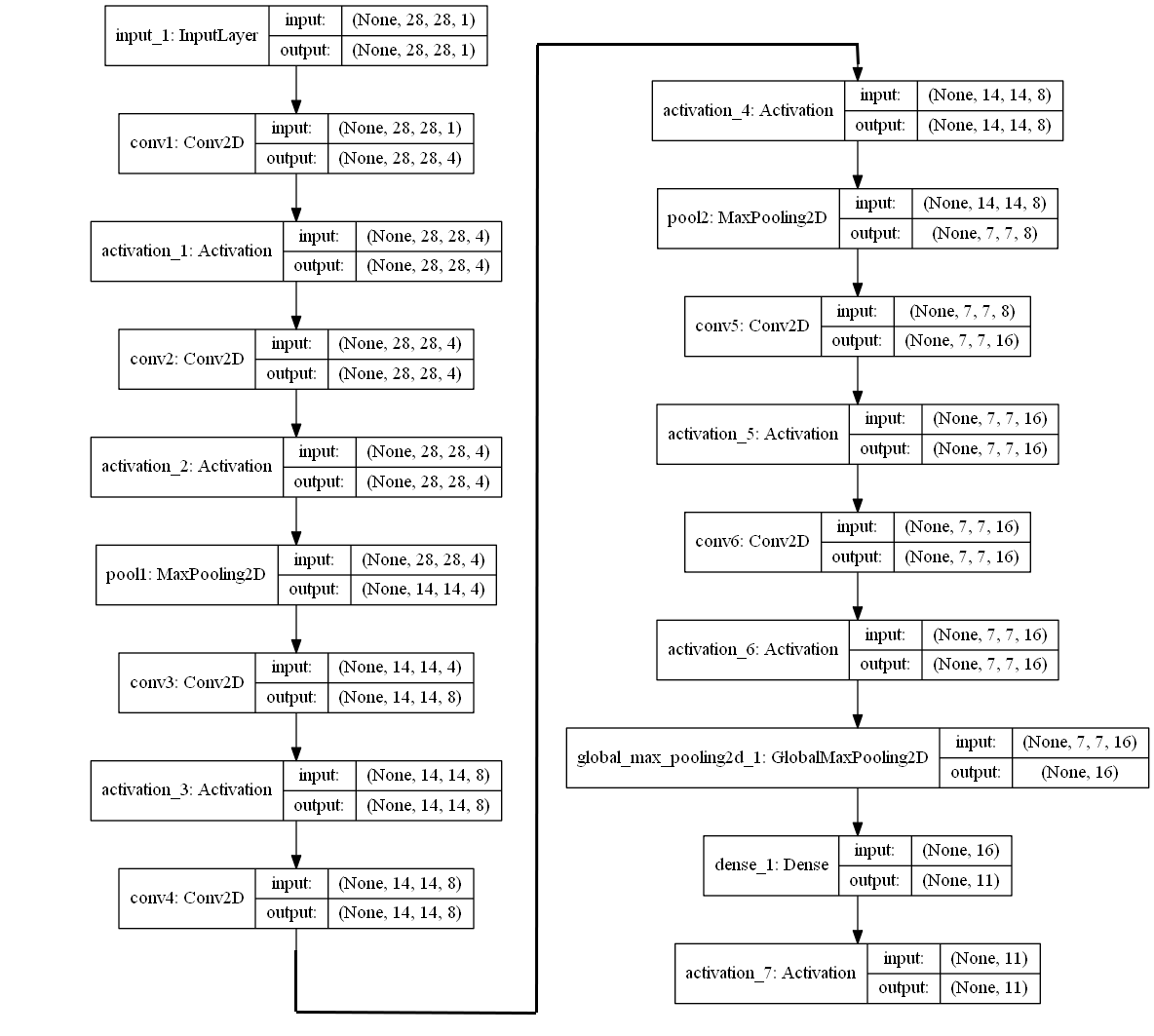}
\caption{Low Weight Digit Detector (LWDD) neural network architecture.}
\label{fig:lwdd}
\end{figure}

\begin{figure}[!t]
\centering
\includegraphics[width=\linewidth]{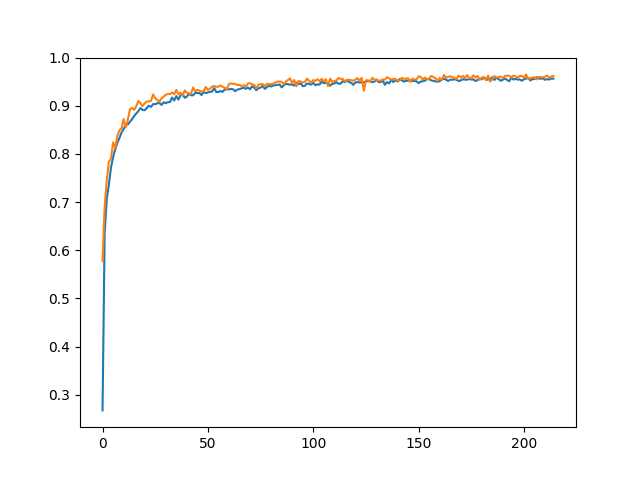}
\caption{Training (orange) and testing (blue) accuracy of the Low Weight Digit Detector network during training on the modified MNIST dataset.}
\label{fig:accuracy}
\end{figure}

In our previous work, we have proposed a VGG Simple neural network (Fig. \ref{fig:vggsimple}) \cite{soloyvev2017fpga}, which is a lightweight modification of the popular VGG architecture \cite{simonyan2014very}. Despite the high performance, the major disadvantage of this model is the number of weights, size of which exceeds FPGA capacity. Besides, the exchange with external memory imposes additional time costs. Moreover, this model involves a "bias" term, which also has to be stored, requires additional processing blocks, and tends to accumulate error if implemented in fixed-point representation. Therefore, we propose a further modification of this architecture that we call Low Weights Digit Detector (LWDD). First, we remove large fully connected layers and bias terms. Then, \texttt{GlobalMaxPooling} layer is added to neural network, instead of \texttt{GlobalAvgPooling}, which is traditionally used, for example, in ResNet50. The efficiency of these layers is approximately the same, while the hardware complexity of finding a maximum is much simpler than mean value calculation from the computational point of view. These introduced changes do not lead to the decrease in network performance. New architecture is shown in Fig. \ref{fig:lwdd}. Changes in neural network structure allows to reduce number of weights from 25,000 to approximately 4,500, and to store all weights in the internal memory of the FPGA.
On the modified MNIST dataset with image augmentations, LWDD neural network achieves ~96\% accuracy (Fig. \ref{fig:accuracy}).

\subsection{Fixed-point calculation implementation}
In neural networks, calculations are traditionally performed with floating point either on GPU (fast) or CPU (slow), for example, using \texttt{float32} type. When implemented at the hardware level, floating-point calculations are slower than fixed-point due to the difficulty of controlling the mantissa and the exponent for various operations.

Let's consider the first convolutional layer of a neural network, which is the main building block of convolutional architectures.

At the layer input is a two-dimensional matrix (original picture) $28\times28$ with values from $[0; 1)$.
It is also known that if $a\in[-1,1]$ and $b\in[-1,1]$, then $a\cdot b\in[-1,1]$.

For $3\times3$ convolution, the value of the certain pixel (i,j) in the second layer can be calculated as following:
\begin{equation}
\label{n}
\begin{aligned}
    n_{i,j}={} & b+w_{00}p_{i-1,j-1}+w_{01}p_{i-1,j}+w_{02}p_{i-1,j+1} + \\
    & w_{10}p_{i,j-1}+w_{11}p_{i,j}+w_{12}p_{i,j+1} + \\
    & w_{20}p_{i+1,j-1} + w_{21}p_{i+1,j}+w_{22}p_{i+1,j+1}.
\end{aligned}
\end{equation}

Since weights $w_{i,j}$ and bias $b$ are known, it is possible to calculate potential minimum $mn$ and maximum $mx$ of the second layer. Let $M=max(|mn|,|mx|)$. If we divide $w_{i,j}$ and $b$ by the value of $M$, we can guarantee that for any configuration of input data, the value on the second layer does not exceed $1$. We call $M$ a reduction coefficient of the layer. For the second layer, we use the same principle, namely, the value at layer input belongs to interval $[-1; 1]$, so we can repeat our reasoning.
For the proposed neural network after all weight reductions to the last layer, the position of the maximum of the last neuron is not changed, that is, the network works equivalently to the neural network without reductions from the point of view of floating-point calculations.

After performing this reduction on each layer, we can move from floating-point calculations to fixed-point calculations, since we know exactly the range of values at each stage of computation. We use the following notation to represent the numbers of bits: $x_b=[x\cdot 2^N]$.

If $z=x+y$, then addition can be expressed as: $z'=x_b+y_b=[x\cdot 2^N]+[y\cdot 2^N]=[(x+y)\cdot 2^N]=[z\cdot 2^N]=z_b$.
Multiplication can be expressed as: $z'=x_b+y_b=[x\cdot 2^N]\cdot [y\cdot 2^N]=[(x\cdot y)\cdot 2^N\cdot 2^N]=[z\cdot 2^N\cdot 2^N]=[z_b\cdot 2^N]$, that is, we have to divide multiplication result by $2^N$ to get the real value, or just shift it by N positions.

If we sort through all possible input images and focus on the potential minimum and maximum values, we can get very large reduction coefficients, such that the accuracy will rapidly decrease from layer to layer. This can require a large width of fixed-point representation of weights and intermediate computational results. To avoid this, we can use all (or a part) of the training set to find most likely maximum and minimum values in each layer. As our experiments are showing, usage of the training set makes it possible to decrease reduction coefficients. At that, we should scale up coefficients by a small margin, either focusing on the value of $3\sigma$ or increasing the maximum by several per cent.

However, under certain conditions, overflow and violation of the calculated range are possible. To address this issue, a hardware implementation requires a detector of such cases and the mechanism for replacement of overflowed values with the maximum for given layer. This can be achieved by minor modifications of a convolutional unit.

For fixed-point calculations with the limited width of weights and intermediate results, rounding errors inevitably arise, accumulate from layer to layer, and can lead to "inaccurate" predictions. We consider "inaccurate" predictions to happen when the predicted value is compared with the prediction by the software implementation, rather then with the true image label. To validate the "accuracy" of predictions, we run all test images through both the floating-point software implementation and fixed-point software implementation (or Verilog benchmark) and then compare predictions. Ratio of mismatches to the total number of tests is a measure of "inaccurate" for the given width of weights and intermediate results. We choose the bit width at which the number of errors is $0$.

When using fixed-point calculations with convolution blocks, two different strategies are possible:
\begin{itemize}
\item rounding after each elementary operation of addition and multiplication;
\item calculation with full accuracy and rounding at the very end of convolution operation.
\end{itemize}

Two experiments are carried out to determine the most effective approach (see Table \ref{table:inaccuracy}). To achieve zero difference from the floating-point model, the number storage requires 17 bits in case of rounding at the beginning, and only 12 bits in case of rounding at the end.

\begin{table}[t!]
\caption{Inaccuracy for different rounding strategies as compared to the software implementation}
\label{table:inaccuracy}
\centering   
\begin{tabular}{|c | c | c |}
\hline
Bit width of weights & Rounding at the beginning & Rounding at the end
 \\
\hline 		
10 & 32.03\% & 1.96\% \\
11 & 15.03\% & 0.65\% \\
12 & 11.76\% & 0.00\% \\
13 & 9.15\% & 0.00\% \\
14 & 3.27\% & 0.00\% \\
15 & 1.96\% & 0.00\% \\
16 & 0.65\% & 0.00\% \\
17 & 0.00\% & 0.00\% \\
18 & 0.00\% & 0.00\% \\
\hline
\end{tabular}
\end{table}
Rounding after each operation slightly increases performance, and significantly increases memory overhead. Therefore, it is advantageous to perform rounding after a convolution block.

\begin{table*}[h!]
\caption{Information on resources used on FPGA after Place \& Route stage}
\label{table:fpga}
\centering   
\begin{tabular}{|c | c | c | c | c |}
\hline
& Logic cells (available: 22320) & 9-bit elements (132) & Internal memory (available: 608256 bit) & PLLs (4) \\
\hline
 Input image converter & 964 (4\%) & 0 (0\%) & 0 (0\%) & 0 (0\%) \\
 Neural network  & 4014 (18\%) & 23 (17\%) & 285428 (47\%) & 0 (0\%) \\
 Weights database  & 0 (0\%) & 0 (0\%) & 70980 (12\%) & 0 (0\%) \\
 Storage of intermediate calculation results  & 1 ($<1\%$) & 0 (0\%) & 214448 (35\%) & 0 (0\%) \\
 Total usage & 5947 (27\%) & 23 (17\%) & 371444 (61\%) & 2 (50\%) \\
\hline
\end{tabular}
\end{table*}

\subsection{FPGA-based hardware implementation}
\begin{figure}[!t]
\centering
\includegraphics[width=\linewidth]{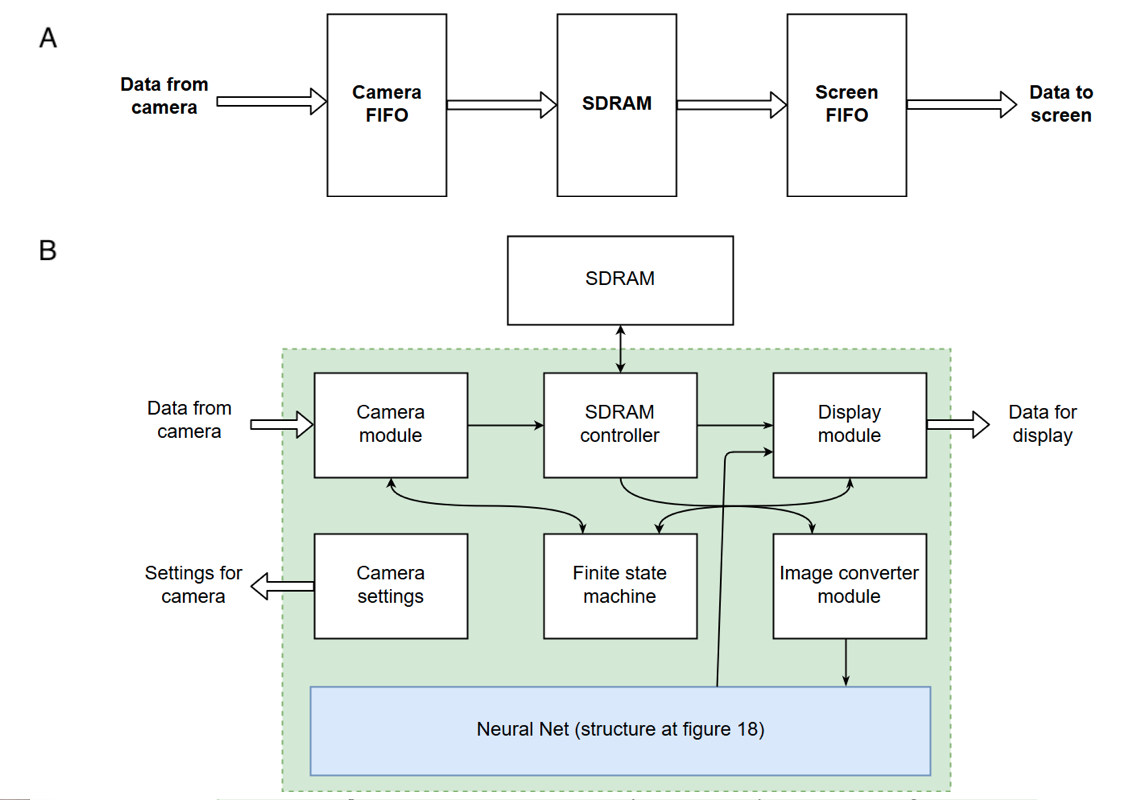}
\caption{Data flow in FPGA implementation: (A) frame data transfer; (B) functional design of the project using Verilog modules.}
\label{fig:dataflow}
\end{figure}

In FPGA-based realization, SDRAM is used to store a video frame from camera. In SDRAM memory on De0-Nano card used in this study, two equal areas for two frames are allocated - current frame is recorded in the first area, and previous frame is read from the other memory area. And after the output is finished, these areas change their roles. When using SDRAM memory in this study, we consider two important issues. First, memory operates at high frequency of 143$MHz$, thus, we face one more problem of transferring of data from the clock domain of camera to the clock domain of SDRAM. Second, in order to achieve maximum speed, writing to SDRAM should be performed by whole transactions, or in "burst". FIFO directly built in FPGA memory is the best way to solve both of these problems. Basic idea is that camera fills FIFO at low frequency, then SDRAM controller reads data at high frequency, and immediately writes them to memory in one transaction.
Data output to TFT screen is organized in the same way. Data from SDRAM are written to screen FIFO, and then are read at the frequency of 10$MHz$. After FIFO has been cleared, the operation is repeated.
Process of frame data transfer through different memory locations is shown in Fig.\ref{fig:dataflow}A, and the functional design of the project using Verilog modules is shown in Fig.\ref{fig:dataflow}B.

\begin{figure}[!t]
\centering
\includegraphics[width=\linewidth]{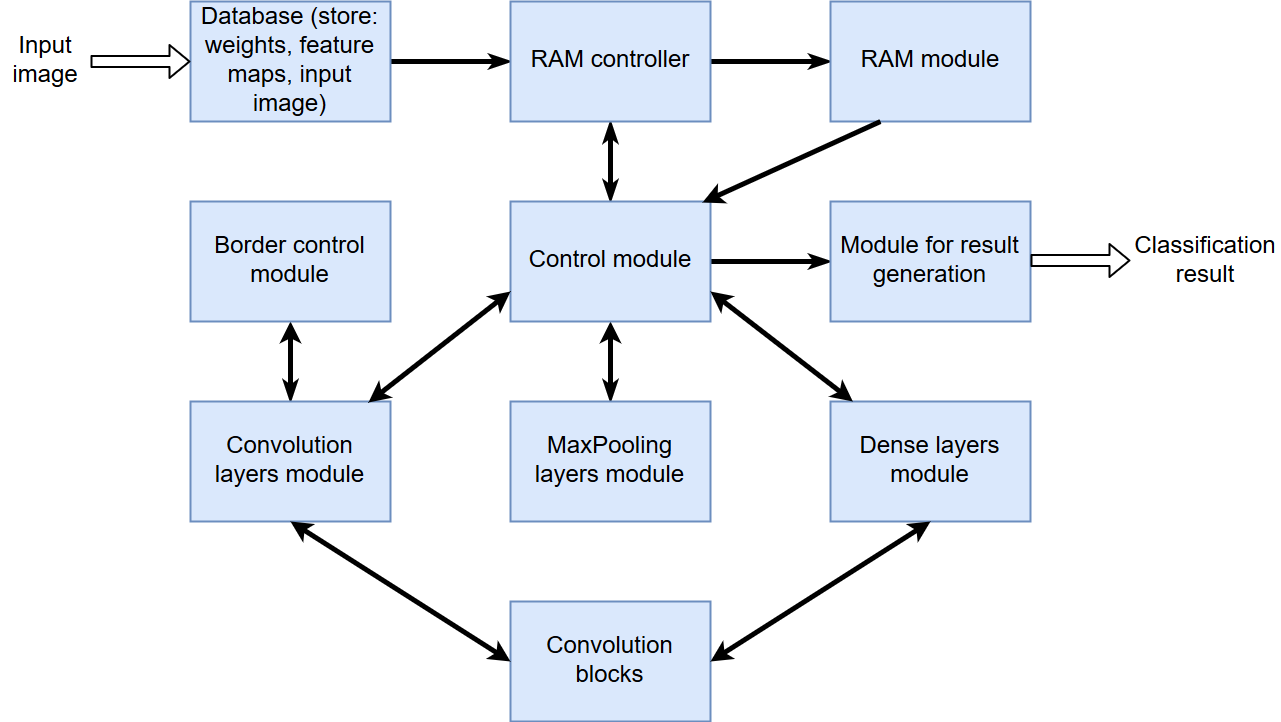}
\caption{Schematic representation of the hardware implementation.}
\label{fig:hardware}
\end{figure}

A picture from the camera, after passing through SDRAM, is displayed on the screen as is, and also is fed to neural network for its recognition through block that converts image to grayscale and decreases resolution. When neural network operation is finished, the result is also output directly to the screen. Since the camera has a large number of operation parameters, they are incorporated in a separate module that uploads them to the camera before the operation starts.
Hardware implementation of the neural network is presented schematically in Fig.\ref{fig:hardware}.

After conversion, an input image is stored in the database, which also stores weight coefficients for each layer that were calculated and wired-in beforehand. As necessary, data from there is downloaded through the controller to the small memory unit for the further use. Everything in neural network is controlled by the main module that keeps the sequence and parameters of network layers. In the hardware realization, not all layers of neural network under test are used; some of them are replaced by other functions. For example, there is no \texttt{ZeroPadding} layer, instead of it module of intermediate image edge detection is applied, which allows to reduce chip memory usage. \texttt{GlobalMaxPooling} layer is replaced by the function from the \texttt{Convolution} layer that immediately gets \texttt{GlobalMaxPooling} layer result by finding the largest value in the intermediate image. The rest of the layers are implemented as separate modules. Since \texttt{Convolution} and \texttt{Dense} layers can use convolutional blocks for calculations, both of them have access to these blocks. Modules contain \texttt{ReLU} activation function, which is used as needed. In the last layer, \texttt{Softmax} activation function is applied. It is implemented as traditional \texttt{Maximum} rather than explicitly, because position of a neuron with the maximum value is always the same for these functions. This modified activation is realized as a separate module, from which results of the neural network operation are received.
To implement the neural network, the specialized \texttt{Convolution} block is used, which performs convolution of $3\times3$ in one clock cycle (we can make such block for other dimensions $4\times4$, $5\times5$, etc., if we have those in the network). This block is a scalar product of vectors and contains 9 multiplications and 8 additions. The same block is used for calculations in the fully connected \texttt{Dense} layer due to splitting the entire set of additions and multiplications into blocks of 9 neurons.

\begin{figure}[!t]
\centering
\includegraphics[width=\linewidth]{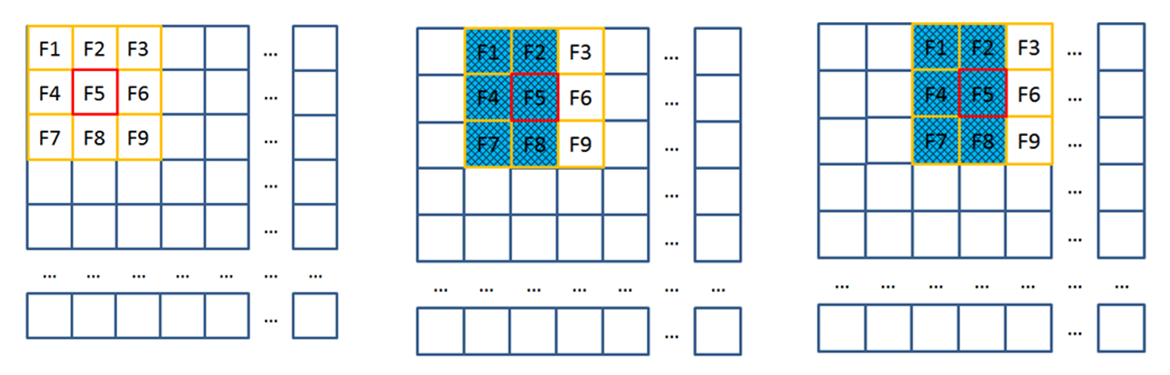}
\caption{Shift register operation: blue indicates data for previous convolution operation obtained at the previous step}
\label{fig:shift}
\end{figure}

\begin{figure}[!t]
\centering
\includegraphics[width=\linewidth]{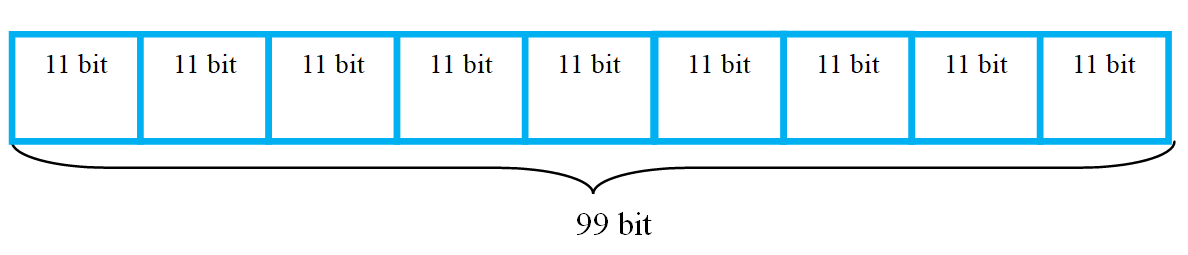}
\caption{Storage of all layer weights as single block.}
\label{fig:storage}
\end{figure}

\begin{table}[h!]
\caption{Clock cycles at each stage of FPGA-based image processing}
\label{table:clock}
\centering   
\begin{tabular}{|c | c |}
\hline
Stage of processing & Number of clock cycles
 \\
\hline 		
Initial image loading & 1570 \\
Loading weights for the 1st conv layer & 76 \\
Processing 1st conv layer & 12605 \\
Loading weights for the 2nd conv layer & 291 \\
Processing 2nd conv layer & 50416 \\
Processing 3rd maxpooling layer & 3164 \\
Loading weights for the 4th conv layer  & 580 \\
Processing 4th conv layer & 25569 \\
Loading weights for the 5th conv layer  & 1155 \\
Processing 5th conv layer & 51136 \\
Processing 6th maxpooling layer & 1623 \\
Loading weights for the 7th conv layer  & 2309 \\
Processing 7th conv layer & 27009 \\
Loading weights for the 8th conv layer & 4611 \\
Processing 8th conv layer (+GlobalMaxPooling) & 54016 \\
Loading weights for the 9th dense layer  & 356 \\
Processing 9th dense layer & 244 \\
Saving result & 16 \\
Total & 236746 \\
\hline
\end{tabular}
\end{table}

\begin{table}[h!]
\caption{Clock cycles per convolution block}
\label{table:cyclesconvo}
\centering   
\begin{tabular}{|c | c | c |}
\hline
&  Clock cycles per 1  frame & Processing speed-up
 \\
\hline 		
1st convolution block & 236746 & - \\
2nd convolution block & 125320 & 1.89 \\
4th convolution block & 67861 & 3.49 \\
\hline
\end{tabular}
\end{table}

\begin{table*}[h!]
\caption{Total resource usage for the project.}
\label{table:total}
\centering   
\begin{tabular}{|c | c | c | c | c | c | c |}
\hline
Weight dimensions & Convolutional blocks & Logical cells & Memory & Embedded Memory 9-bit elements & Critical path delay & Max. FPS
 \\
\hline 		
\multirow{4}{*}{11 bit} & 1 & 3750 & 232111 & 25 & 21.840 & 193 \\
						& 2 & 4710 & 309727 & 41 & 22.628 & 352 \\
						& 4 & 6711 & 464959 & 77 & 23.548 & 625 \\
\hline 
\multirow{4}{*}{12 bit} & 1 & 3876 & 253212 & 25 & 24.181 & 174 \\
						& 2 & 4905 & 337884 & 41 & 24.348 & 327 \\
						& 4 & 10064 & 589148 & 77 & - & - \\
\hline 
\multirow{4}{*}{13 bit} & 1 & 3994 & 274313 & 25 & 22.999 & 183 \\
						& 2 & 5124 & 366041 & 41 & 25.044 & 318 \\
						& 4 & 8437 & 549497 & 77 & - & - \\
\hline
\end{tabular}
\end{table*}

\subsection{Additional optimization of calculations}
To increase the performance, a number of techniques are applied that made it possible to reduce the number of cycles required for one image classification.

\subsubsection{Increasing of convolution blocks number}
If there is enough free space in FPGA, we can improve the performance by increasing the number of convolution blocks, thereby multiplying productivity. Consider the second convolutional block in the proposed neural network LWDD. There are 4 of $28\times28$ images at the layer input, and 16 blocks of weights are given. To calculate the set of outputs for this layer, also consisting of 4 images, we have to perform four multiplications of the same set of pixels by different sets of weights. If there is only one convolutional block, this takes at least 4 cycles, but if there are 4 such blocks, then only one clock cycle is needed, thus \texttt{Convolution} layer calculation speeds up 4 times.

\subsubsection{Shift register}
To perform an elementary convolution operation, we have to get values of 9 neighboring pixels from an input image, then next 9 pixels, 6 of which have already been received in the previous step (see Fig. \ref{fig:shift}).
To shorten the time for the necessary data call up, shift register is developed to keep new data at their input and at the same time to "push out" old data. Thus, each step requires only 3 new values instead of 9.

\subsubsection{Storing of all data for one Convolution operation at the same address}
When we call up data that are necessary for calculations, one clock cycle is used for each value. Therefore, in order to reduce time spent on downloading required data, as well as convenience of access, prior to putting into internal memory of FPGA, data are stacked in blocks of 9 pieces, after which they are accessible at one address. With such memory arrangement, we can perform the extraction of weights in one clock cycle and, thus, speed up calculations for convolutional and fully connected layers. Example is shown in Fig. \ref{fig:storage}.

\section{Performance results}
The proposed design is successfully implemented in FPGA. Details on logic cells number and memory usage are given in Table \ref{table:fpga}. These numerical results demonstrate low hardware requirements of the proposed model architecture. Moreover, for this implementation, the depth of the neural network can be further increased without exhausting the resources of this specific hardware.

In this implementation, input images are processed in real time, and the original image is displayed along with the result. Classification of one image requires about 230 thousand clock cycles and we achieve overall processing speed with the large margin over 150 frames/sec. The detailed distribution of clock cycles at each processing step is given in Table \ref{table:clock}.

If performance is insufficient and spare logic cells are available, we can speed up calculations by adding convolutional blocks that perform computations in parallel, as suggested in \cite{soloyvev2017fpga}. Table \ref{table:cyclesconvo} shows number of clock cycles required to process one frame using different number of convolutional blocks. Table \ref{table:total} shows total resources required for entire FPGA-based project implementation for different weight dimensions and different number of convolution blocks. Missing values denotes the cases that Quartus could not synthesize due to the lack of FPGA resources.

Video demonstrating the real-time handwritten digit classification from the mobile camera video feed is available on YouTube \cite{fpga2018demo}. Source code for both software and hardware implementations is available on GitHub \cite{fpga2018code}.

\section{Conclusions}
In this work we propose a design and an implementation of FPGA-based CNN with fixed-point calculations that allows to achieve the exact performance of the corresponding software implementation on the live handwritten digit recognition problem. Due to the reduced number of parameters we avoid common issues with memory bandwidth. Suggested method can be implemented on a very basic FPGAs, but also is scalable for the use on FPGAs with large number of logical cells. Additionally, we demonstrate how existing open datasets can be modified in order to better adapt them for real-life applications. Finally, in order to promote the reproducibility of results, facilitate open-scientific development, and enable collaborative validation, source code, documentation, and all results from this study are made available online.

There are many possible ways to improve performance of a hardware implementations of neural networks. While we explored and implemented some of them in this work, only relatively shallow deep neural networks were considered, without additional architectural features, such as skip-connections. Implementing even deeper networks with multiple dozens of layers is problematic, since all layer weights would not fit into the FPGA memory and will require the use of the external RAM, which can lead to the decrease in performance. Moreover, due to the large number of layers, error accumulation will increase and will require wider bit range to store fixed-point weight values. In the future, we plan to consider implementing on FPGA specialized lightweight neural network architectures that are currently successfully used on mobile devices. This will allow to use the same hardware implementation for different tasks by fine-tuning the architecture using pre-trained weights.


%



\section*{Acknowledgment}
Research has been conducted with the financial support from the Russian Science Foundation (grant 17-19-01645).


\ifCLASSOPTIONcaptionsoff
  \newpage
\fi



\bibliographystyle{IEEEtran}
\bibliography{./bibtex/bib/IEEEabrv,./bibtex/bib/paper}
\end{document}